\documentclass[runningheads]{llncs}

\usepackage{eccvabbrv}

\usepackage{graphicx}
\usepackage{booktabs}
\usepackage{multirow}
\usepackage{cite}

\usepackage[accsupp]{axessibility}  

\usepackage{orcidlink}

\begin{document}

\title{RGMIM: Region-Guided Masked Image Modeling for Learning Meaningful Representations from X-Ray Images} 

\titlerunning{Region-Guided Masked Image Modeling}

\authorrunning{G.~Li et al.}

\author{Guang Li\orcidlink{0000-0003-2898-2504} \and Ren Togo\orcidlink{0000-0002-4474-3995} \and Takahiro Ogawa\orcidlink{0000-0001-5332-8112} \and Miki Haseyama\orcidlink{0000-0003-1496-1761}}

\institute{Hokkaido University\\
\email{\{guang,togo,ogawa,mhaseyama\}@lmd.ist.hokudai.ac.jp}}

\maketitle

\begin{abstract}
  In this study, we propose a novel method called region-guided masked image modeling (RGMIM) for learning meaningful representations from X-ray images. Our method adopts a new masking strategy that utilizes organ mask information to identify valid regions for learning more meaningful representations. We conduct quantitative evaluations on an open lung X-ray image dataset as well as masking ratio hyperparameter studies. When using the entire training set, RGMIM outperformed other comparable methods, achieving a 0.962 lung disease detection accuracy. Specifically, RGMIM significantly improved performance in small data volumes, such as 5\% and 10\% of the training set compared to other methods. RGMIM can mask more valid regions, facilitating the learning of discriminative representations and the subsequent high-accuracy lung disease detection. RGMIM outperforms other state-of-the-art self-supervised learning methods in experiments, particularly when limited training data is used.
\end{abstract}

\section{Background}
\begin{figure*}[t]
        \centering
        \includegraphics[width=12.0cm]{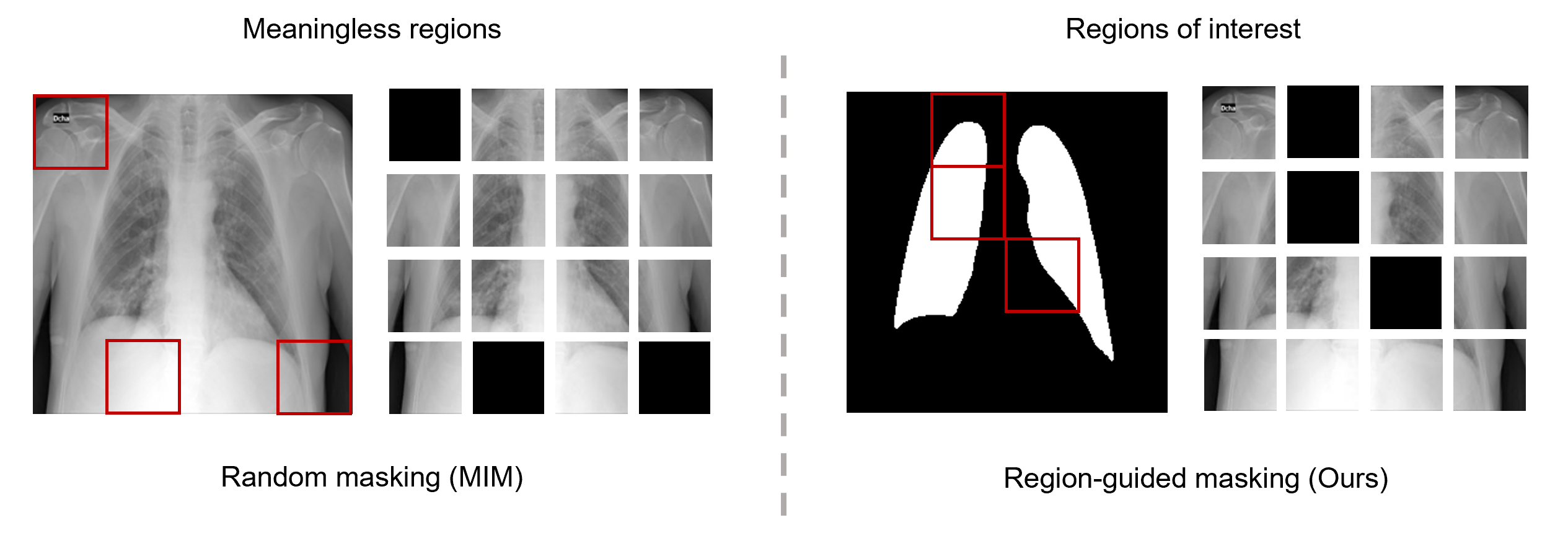}
        \caption{Comparison of the random masking and the proposed region-guided masking.}
        \label{fig0}
\end{figure*}
Self-supervised learning is regarded as the most essential part of artificial intelligence and has made great progress in the medical field~\cite{chen2019self, krishnan2022self, shurrab2022self}.
Early self-supervised learning techniques usually learn representations through a simple prior task that applies transformations such as rotate~\cite{gidaris2018unsupervised} and jigsaw puzzle~\cite{noroozi2016unsupervised} to the input image and make the model predict the properties of the transformation from the transformed image.
Afterward, contrastive learning methods have been demonstrated to be efficient on different tasks~\cite{liu2021self}.
These approaches aim to make the distances between various view representations enhanced from the same image (i.e., positive sample pairs) closer while making the distances between view representations enhanced from different images (i.e., negative sample pairs) farther~\cite{chen2020simple}.
Nevertheless, as these techniques depend on data augmentation, the hyperparameters and data augmentation strategies need to be changed frequently~\cite{li2022tri}.
\par
As one of the hot research topics of self-supervised learning, masked image modeling (MIM) has recently attracted more and more attention~\cite{bao2022beit, he2022masked, xie2022simmim}.
In the discipline of natural language processing, masked language modeling (MLM) is a task that masks a portion of the input words and attempts to predict the masked words~\cite{devlin2019bert}.
Accordingly, recovering a large number of masked patches from a small number of unmasked patches allows the model to learn sufficient semantic information~\cite{wei2022masked}.
In comparison to contrastive learning techniques relying on significant data augmentation, MIM is also more straightforward and reliable.
As a result, it is valuable to investigate the role of MIM on real-world datasets such as medical image datasets~\cite{chen2023masked}.
\par
Traditional MIM techniques frequently employ a random masking strategy for ordinary images~\cite{he2022masked}.
Medical images, in comparison to ordinary images, often exhibit a smaller region of interest for disease classification~\cite{shin2016deep}.
For instance, when applied to lung disease classification, the regions outside the lung may not have the information needed to make a judgment, which could prevent the random masking strategy from learning enough information.
\par
In this paper, we propose a novel method called region-guided masked image modeling (RGMIM) for learning meaningful representations from X-Ray images. 
Our approach is designed to address the issue that medical images often have a smaller region of interest compared to ordinary images, by using organ mask information to identify valid regions for learning more useful information. 
Specifically, we introduce a new masking strategy in our method.
As depicted in Fig.~\ref{fig0}, the proposed region-guided masking strategy can resolve the shortcomings of random masking in medical images.
RGMIM obtained a 0.962 lung disease classification accuracy on an open X-ray image dataset according to experimental results.
When limited training data was used, RGMIM drastically outperformed other state-of-the-art (SOTA) self-supervised learning methods significantly.
Our main contributions are listed as follows:
\begin{itemize}
    \item We propose a novel RGMIM for learning meaningful representations from X-ray images.
    \item RGMIM significantly outperforms other SOTA self-supervised learning methods on an open X-ray dataset, especially when using limited training data.
\end{itemize}
\section{Methods}
\begin{figure*}[t]
        \centering
        \includegraphics[width=12cm]{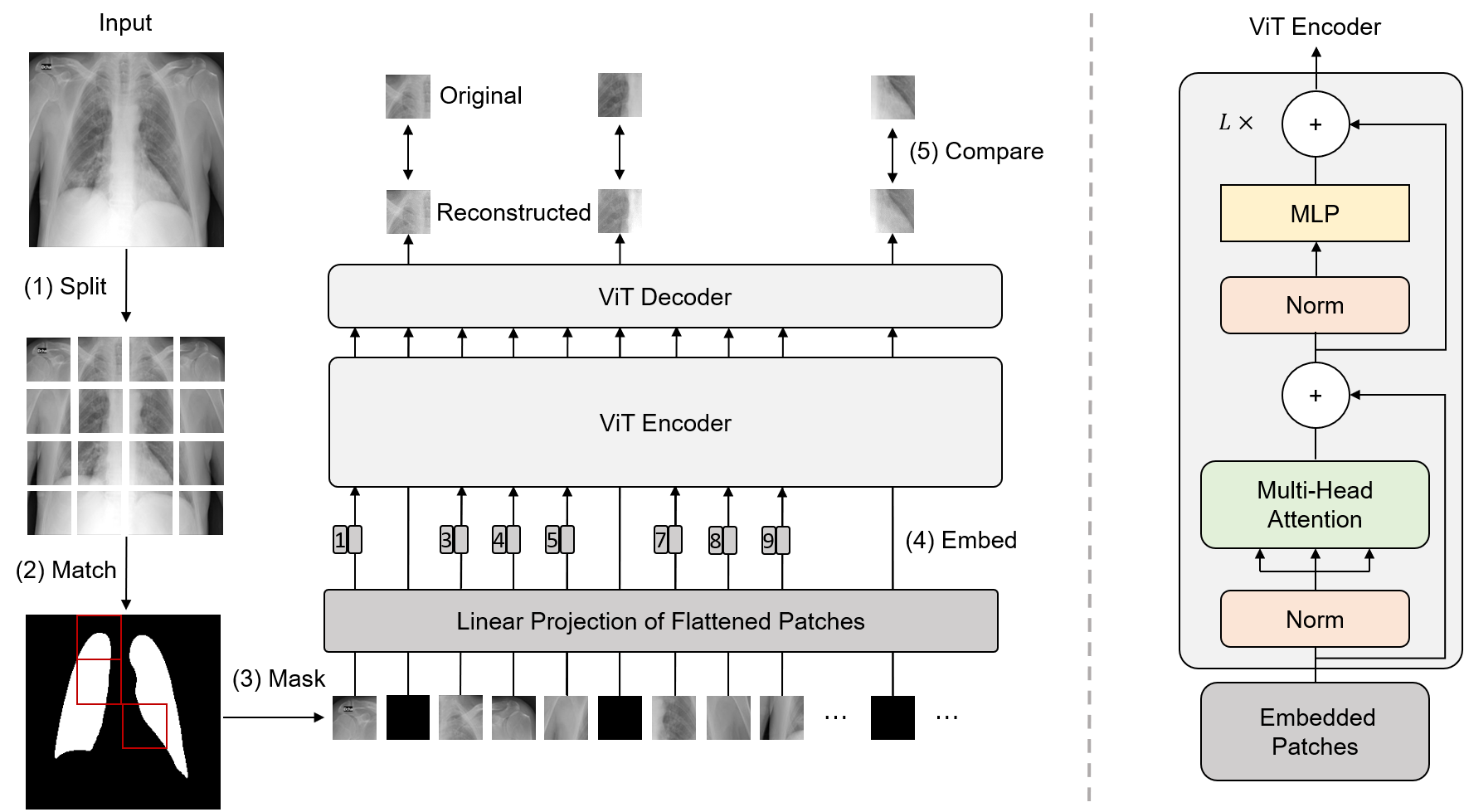}
        \caption{Overview of RGMIM. The left indicates the pipeline and the right show the structure of the ViT encoder.}
        \label{fig1}
\end{figure*}
RGMIM is a pretraining task that uses Vision Transformer (ViT) to compare the original patches and the reconstructed masked patches~\cite{dosovitskiy2020vit}.
The network structure contains a ViT encoder and a ViT decoder.
The ViT encoder transforms the unmasked lung X-ray patches into a latent representation.
From the latent representation, the ViT decoder can reconstruct masked lung X-ray patches.
An overview of RGMIM is shown in Fig.~\ref{fig1}.
The left indicates the pipeline and the right show the structure of the ViT encoder. 
RGMIM consists of five steps: (1) splitting the input lung X-ray image into patches, (2) matching the patches with the lung mask image, (3) masking some valid patches, (4) embedding the unmasked patches, and (5) compare the reconstructed patches and masked patches.
In the following subsections, we go over the region-guided masking strategy, ViT encoder and decoder, pretraining process, fine-tuning, and lung disease classification in detail.
\subsection{Region-guided masking strategy}
First, given an input lung X-ray image $X \in \mathbb{R}^{H\times W\times C}$, we divide it into $n$ patches $X_{N} \in \mathbb{R}^{n\times (T^{2}\times C)}$, where ($H, W$) denotes the size of the original lung X-ray image, $C$ denotes the number of channels, ($T, T$) denotes the size of splitted patches, and $n = HW/T^{2}$ is the number of patches.
For ordinary images, traditional MIM methods frequently employ a random masking strategy.
However, lung disease classification is only related to lung characteristics, and information from areas other than the lungs is meaningless.
As shown on the left of Fig.~\ref{fig0}, if we use a random masking strategy for lung X-ray images, we may mask too many ineffective areas (e.g., areas outside the lungs), and thus the self-supervised learning process cannot learn sufficient representations.   
\par
To tackle the aforementioned challenge, a novel masking strategy that utilizes lung mask information to identify the valid regions for learning more informative features for lung disease classification is proposed in this study.
For example, for each input lung X-ray image $X$, there is a corresponding lung mask image that represents the lung region.
And we match the split patches $X_{N}$ with the lung mask image.
We define split patches as valid lung X-ray patches if they overlap with the lung mask image.  
Then we randomly mask $m = n \times \sigma$ valid lung X-ray patches $X_{M}$ with the masking ratio $\sigma$ and obtain the remaining $u = n \times (1-\sigma)$ unmasked patches $X_{U}$.
The unmasked patches $X_{U}$ served as the effective input sequence of the ViT encoder.
With the proposed region-guided masking strategy, we can mask more valid lung-related regions, facilitating the learning of discriminative representations and the subsequent high-accuracy lung disease classification. 
\subsection{ViT encoder and decoder}
Following that, because all layers of the ViT encoder use a constant latent vector size $D$, we flatten the $u$ unmasked patches and transform them to $D$ dimensions using a trainable linear projection, as shown below:
\begin{equation}
\mathbf{z}_{0} = [X^{1}_{U}\mathbf{E}; X^{2}_{U}\mathbf{E};\dotsb; X^{u}_{U}\mathbf{E}] + \mathbf{E}_\mathrm{pos}, 
\end{equation}
where $\mathbf{E} \in \mathbb{R}^{((T^{2}\times C) \times D)}$ denotes the patch embeddings and $\mathbf{E}_\mathrm{pos} \in \mathbb{R}^{(u \times D)}$ denotes the position embeddings.
Position embeddings $\mathbf{E}_\mathrm{pos}$ are appended to the patch embeddings $\mathbf{E}$ to preserve positional information. 
We employ learnable one-dimensional position embeddings, and the obtained sequence of embedding vectors $\mathbf{z}_{0}$ is input to the ViT encoder.
The ViT encoder has a total number of $L$ blocks.
Each block is made up of a multi-head self-attention (MSA) module, as well as a multilayer perceptron (MLP), layer normalization (LN), and residual connections.
Each layer of the MLP has a Gaussian error linear unit activation function.
The global average pooling (GAP) is used to compute the final output embeddings $\mathbf{y}$ of unmasked patches $X_{U}$.
The embedding process is defined as follows:
\begin{equation}
\hat{\mathbf{z}}_{l} = \mathrm{MSA}(\mathrm{LN}(\mathbf{z}_{l-1})), \quad l = 1 ... L
\end{equation}
\begin{equation}
\mathbf{z}_{l} = \mathrm{MLP}(\mathrm{LN}(\hat{\mathbf{z}}_{l})), \quad l = 1 ... L
\end{equation}
\begin{equation}
\mathbf{y} = \mathrm{GAP}(\mathrm{LN}(\mathbf{z}_{L})). 
\end{equation}
where $\hat{\mathbf{z}}_{l}$ and $\mathbf{z}_{l}$ denote sequences of embedding vectors in intermediate layers.
\par
The ViT decoder is an inverse version of the ViT encoder. 
The inputs of the ViT decoder are final output embeddings $\mathbf{y}$ and masked tokens.
The ViT decoder is only used in the pretraining process and can be designed differently from the ViT encoder.
The representations learned by the MIM are also well when using a lightweight ViT decoder with fewer layers, and it can significantly decrease the network parameters and pretraining time~\cite{he2022masked}.
%
%
\subsection{Pretraining process}
Finally, by predicting the pixel values of masked patches, and RGMIM reconstructs the original patches $X_{M}$.
Each element in the ViT decoder output is a vector of pixel values representing a patch.
Following the ViT decoder, a linear projection layer with the number of output channels equal to the number of pixel values in a patch is added.
And the ViT decoder’s outputs are reshaped to generate reconstructed patches $Y_{M}$.
We calculate the $L_{2}$ error between the reconstructed patches $Y_{M}$ and original patches $X_{M}$ in the pixel-level as follows:
\begin{equation}
\mathcal{L} = \frac{1}{m \times T^{2}\times C} || Y_{M} - X_{M} ||^{2}_{2},
\end{equation}
where $m \times T^{2}\times C$ denotes the number of elements in original patches.
As a self-supervised pretraining process, the ViT encoder can learn discriminative representations from non-labeled lung X-ray images.
\subsection{Fine-tuning and lung disease classification}
The ViT decoder is discarded following the RGMIM pretraining process, and the pretrained ViT encoder with a linear classification layer is used to fine-tune labeled lung X-ray images using the cross-entropy loss.
Given unknown lung X-ray images, the fine-tuned ViT encoder with a linear classification layer can predict the categories of the input lung X-ray images and perform lung disease classification during the test phase. 
Since the ViT encoder has learned discriminative representations from lung X-ray images, the fine-tuning process can be important shortened and still achieve great classification accuracy with limited labeled lung X-ray images.
\section{Experiments}
\begin{table}[t]
	\begin{center}
		\caption{Hyperparameters of the proposed method.}
		\label{tab0}
		\begin{tabular}[t]{lc}
		\hline
		Hyperparameter  & Value \\
		\hline\hline
            Pretraining epoch & 40 \\
            Fine-tuning epoch & 30 \\ 
            Batch size    & 256 \\
		Base Learning rate & 1.5e-4 \\
		Weight decay  & 0.05 \\
      	Momentum $\beta_{1}$ & 0.9 \\
            Momentum $\beta_{2}$ & 0.95 \\
		Patch size $T$ & 16 \\
   	  Masking ratio $\sigma$ & 75\% \\
		\hline
		\end{tabular}
	\end{center}
\end{table}
In this section, we perform extensive experiments and hyperparameter studies to validate RGMIM's effectiveness.
The dataset and settings are depicted in subsection 3.1.
The findings and analysis are presented in subsection 3.2.
\subsection{Dataset and Settings}
The dataset used in our study is an open lung X-ray image dataset~\cite{rahman2021exploring}.
Lung X-ray images were gathered from a variety of publicly available datasets, online sources, and published articles. 
Each lung X-ray image has a corresponding lung mask image.
The lung mask images are confirmed by expert radiologists as ground truth.
The open dataset has 21,165 lung X-ray images.
The dataset has four categories, including COVID-19, Lung Opacity, Normal, and Viral Pneumonia.
We randomly choose 80\% of the lung X-ray images in each category as the training set (i.e., 16,933 images) and the remaining 20\% as the test set (i.e., 4,232 images).
All lung X-ray images are gray-scale and are resized to 224 $\times$ 224 pixels.
All lung X-ray images without label data in the training set were employed for the pretraining process.
The experiments were performed on an NVIDIA RTX A6000 GPU with 48G memory and employed the PyTorch framework.
In our method, pre-training and fine-tuning take about 53 and 21 minutes, respectively.
\par
As comparison methods, we used five SOTA self-supervised learning methods as comparative methods, including masked autoencoder (MAE)~\cite{he2022masked}, self-knowledge distillation based self-supervised learning (SKD)~\cite{li2022self}, cross-view self-supervised learning (Cross)~\cite{li2022covid}, bootstrap your own latent (BYOL)~\cite{grill2020bootstrap}, and simple siamese self-supervised learning (SimSiam)~\cite{chen2021exploring}.
As baseline methods, Transfer learning (fine-tuning from ImageNet~\cite{deng2009imagenet}) and training from scratch were used, i.e., Transfer and From Scratch.
For RGMIM and MAE, we employed the same settings in all experiments, except for the masking strategy.
For example, we employed ViT-Base~\cite{dosovitskiy2020vit} as the ViT encoder and AdamW~\cite{loshchilov2019decoupled} as the optimizer.
ViT-Base is a small ViT model with layer $L = 12$, latent vector size $D = 768$, 12 heads, and 3072 MLP, which often corresponds to ResNet-50~\cite{he2016deep}.
The ViT decoder employed in our experiments has layer $L = 8$, latent vector size $D = 512$, and the number of heads is 16.
More details about hyperparameters are depicted in Table~\ref{tab0}.
\par
For SKD, Cross, BYOL, and SimSiam, we employed ResNet-50 as the backbone network.
All of these contrastive learning methods make use of standard data augmentations (e.g., flipping, resizing, cropping, Gaussian blur).
For Transfer and From Scratch, we used both ViT-Base and ResNet-50.
To test RGMIM's effectiveness in different training data volumes, we also randomly selected data at 1\%, 5\%, 10\%, and 50\% of the total dataset size for fine-tuning, with the ratio selected in each category being the same.
Because the information density of the image is lower than language, different from the MLM employing a low masking ratio (e.g., 0.15), we set the masking ratio $\sigma$ to 0.75 for lung X-ray images.
We also studied the masking ratio for RGMIM and MAE using hyperparameters.
We used a four-class accuracy as the evaluation metric.
\begin{table*}[t]
    \centering
    \caption{lung disease classification accuracy of different fine-tuning data volumes.}
    \label{tab1}
    \begin{tabular}{lcccccccc}
    \hline
    Method & Architecture & 1\% & 5\% & 10\% & 50\% &100\% \\\hline\hline
    RGMIM & ViT-Base 
    & \bfseries{0.771} & \bfseries{0.893} & \bfseries{0.919} & \bfseries{0.957} & \bfseries{0.962} \\
    MAE~\cite{he2022masked} & ViT-Base
    & 0.754 & 0.875 & 0.903 & 0.948 & 0.956 \\
    SKD~\cite{li2022self} & ResNet-50
    & 0.742 & 0.812 & 0.896 & 0.947 & 0.957 \\
    Cross~\cite{li2022covid} & ResNet-50
    & 0.747 & 0.795 & 0.817 & 0.934 & 0.953 \\
    BYOL~\cite{grill2020bootstrap} & ResNet-50
    & 0.683 & 0.754 & 0.790 & 0.933 & 0.954 \\
    SimSiam~\cite{chen2021exploring} & ResNet-50
    & 0.623 & 0.700 & 0.781 & 0.929 & 0.949 \\
    Transfer & ViT-Base
    & 0.689 & 0.861 & 0.893 & 0.940 & 0.953 \\
    Transfer & ResNet-50
    & 0.539 & 0.619 & 0.665 & 0.913 & 0.936 \\
    From Scratch & ViT-Base
    & 0.413 & 0.580 & 0.645 & 0.810 & 0.848 \\
    From Scratch & ResNet-50
    & 0.284 & 0.496 & 0.532 & 0.619 & 0.774 \\
    \hline
    \end{tabular}
\end{table*}
\begin{table}[t]
    \centering
    \caption{Hyperparameter studies on masking ratio.}
    \label{tab2}
    \begin{tabular}{lccc}
    \hline
    Method & Masking ratio & Accuracy \\\hline\hline
    \multirow{6}*{RGMIM} & 0.15 & 0.949 \\
    & 0.30 & 0.953 \\
    & 0.45 & 0.955 \\
    & 0.60 & 0.958 \\
    & 0.75 &\bfseries{0.962} \\
    & 0.90 & 0.957 \\\hline
    \multirow{6}*{MAE~\cite{he2022masked}} & 0.15 & 0.936 \\
    & 0.30 & 0.942 \\
    & 0.45 & 0.947 \\
    & 0.60 & 0.951 \\
    & 0.75 & \bfseries{0.956} \\
    & 0.90 & 0.953 \\\hline
    \end{tabular}
\end{table}
\subsection{Experimental Results}
The lung disease classification accuracy in different training data volumes is displayed in Table~\ref{tab1}.
Table~\ref{tab1} shows that RGMIM outperformed other comparative methods, achieving 0.962 classification accuracy when using the entire training set.
Specifically, compared to other methods, RGMIM significantly improved lung disease classification in small data volumes, such as 5\% and 10\% of the training set (846 and 1,693 images), and achieved 0.957 classification accuracy even for 50\% of the training set.
We can see from Table~\ref{tab1} that the knowledge learned by RGMIM is useful for self-supervised learning on lung X-ray images and improves representation learning performance for lung disease classification.
Table~\ref{tab2} reveals the lung disease classification accuracy of RGMIM and MAE when employing different masking ratios.
From Table~\ref{tab2}, when the masking ratio $\sigma = 0.75$, the best lung disease classification accuracy is achieved, indicating that the proper masking ratio can improve the classification accuracy of both methods.
In addition, we observe that RGMIM outperforms MAE in terms of robustness, especially when the masking ratio is relatively low, demonstrating the superiority of our method in handling incomplete lung X-ray images.
\begin{figure}[t]
        \centering
        \includegraphics[width=7.0cm]{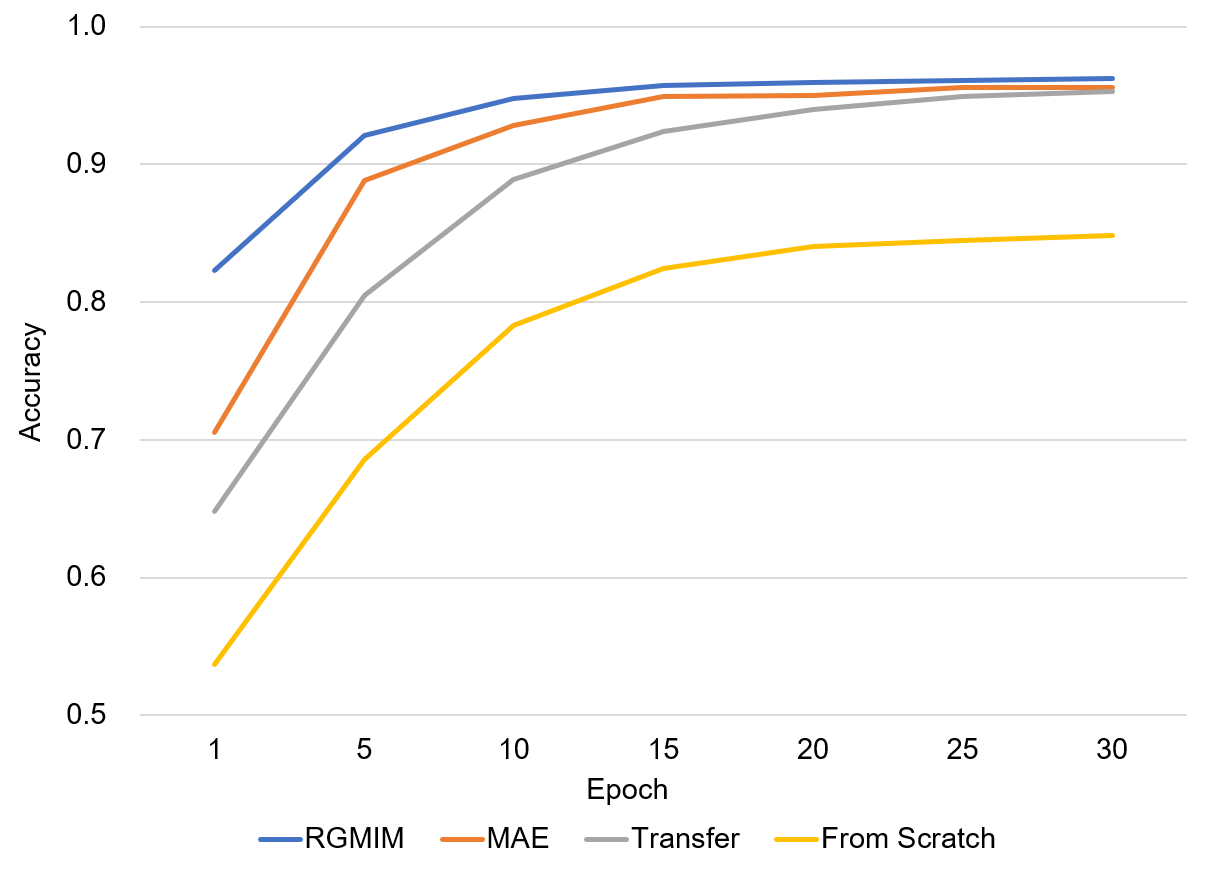}
        \caption{lung disease classification accuracy as fine-tuning epoch number increases. All methods use the ViT-Base model.}
        \label{fig3}
\end{figure}
%
%
In Fig.~\ref{fig3}, we present the accuracy changes of RGMIM when using all lung X-ray images for fine-tuning. 
The results indicate that RGMIM exhibits superior accuracy and faster convergence speed compared to MAE. 
Specifically, after only ten epochs of learning, the classification accuracy of RGMIM has already begun to converge, while MAE is still in the process of convergence. 
The observed difference in convergence speed between RGMIM and MAE can be attributed to the fact that RGMIM incorporates the spatial information of lung X-ray images through the region-guided masking strategy, which helps it to focus on the most informative regions of the images during training. 
Consequently, RGMIM can achieve higher accuracy with fewer iterations, making it an efficient and practical solution for lung disease classification from lung X-ray images.
\section{Conclusion}
This paper presents region-guided masked image modeling (RGMIM), a novel self-supervised learning method for lung disease classification using X-rays. The main contribution of RGMIM is a new masking strategy that utilizes lung mask data to focus on relevant areas, improving the model's ability to learn useful information for disease classification. RGMIM achieved a 0.962 classification accuracy on an open lung X-ray dataset, demonstrating its effectiveness. Notably, RGMIM outperforms other state-of-the-art self-supervised learning methods, especially when dealing with limited training data. This suggests its potential to efficiently handle real-world challenges in medical imaging.

\section*{Acknowledgements}
This study was supported by JSPS KAKENHI Grant Numbers JP23K21676, JP23K11141, and JP24K23849.

%
%
\bibliographystyle{splncs04}
\bibliography{main}

\begin{thebibliography}{10}
\providecommand{\url}[1]{\texttt{#1}}
\providecommand{\urlprefix}{URL }
\providecommand{\doi}[1]{https://doi.org/#1}

\bibitem{bao2022beit}
Bao, H., Dong, L., Wei, F.: Beit: Bert pre-training of image transformers. Proceedings of the International Conference on Learning Representations (ICLR) (2022)

\bibitem{chen2019self}
Chen, L., Bentley, P., Mori, K., Misawa, K., Fujiwara, M., Rueckert, D.: Self-supervised learning for medical image analysis using image context restoration. Medical Image Analysis  \textbf{58},  101539 (2019)

\bibitem{chen2020simple}
Chen, T., Kornblith, S., Norouzi, M., Hinton, G.: A simple framework for contrastive learning of visual representations. Proceedings of the International Conference on Machine Learning (ICML) (2020)

\bibitem{chen2021exploring}
Chen, X., He, K.: Exploring simple siamese representation learning. pp. 15750--15758. Proceedings of the IEEE/CVF Conference on Computer Vision and Pattern Recognition (CVPR) (2021)

\bibitem{chen2023masked}
Chen, Z., Agarwal, D., Aggarwal, K., Safta, W., Balan, M.M., Brown, K.: Masked image modeling advances 3d medical image analysis. pp. 1969--1979. Proceedings of the IEEE/CVF Winter Conference on Applications of Computer Vision (WACV) (2023)

\bibitem{deng2009imagenet}
Deng, J., Dong, W., Socher, R., Li, L.J., Li, K., Fei-Fei, L.: Imagenet: A large-scale hierarchical image database. pp. 248--255. Proceedings of the IEEE Conference on Computer Vision and Pattern Recognition (CVPR) (2009)

\bibitem{devlin2019bert}
Devlin, J., Chang, M.W., Lee, K., Toutanova, K.: Bert: Pre-training of deep bidirectional transformers for language understanding. Proceedings of the Annual Conference of the North American Chapter of the Association for Computational Linguistics (NAACL) (2019)

\bibitem{dosovitskiy2020vit}
Dosovitskiy, A., Beyer, L., Kolesnikov, A., Weissenborn, D., Zhai, X., Unterthiner, T., Dehghani, M., Minderer, M., Heigold, G., Gelly, S., Uszkoreit, J., Houlsby, N.: An image is worth 16x16 words: Transformers for image recognition at scale. Proceedings of the International Conference on Learning Representations (ICLR) (2021)

\bibitem{gidaris2018unsupervised}
Gidaris, S., Singh, P., Komodakis, N.: Unsupervised representation learning by predicting image rotations. Proceedings of the International Conference on Learning Representations (ICLR) (2018)

\bibitem{grill2020bootstrap}
Grill, J.B., Strub, F., Altch{\'e}, F., Tallec, C., Richemond, P., Buchatskaya, E., Doersch, C., Avila~Pires, B., Guo, Z., Gheshlaghi~Azar, M., Piot, B., Kavukcuoglu, K., Munos, R., Valko, M.: Bootstrap your own latent-a new approach to self-supervised learning. pp. 21271--21284. Proceedings of the Advances in Neural Information Processing Systems (NeurIPS) (2020)

\bibitem{he2022masked}
He, K., Chen, X., Xie, S., Li, Y., Doll{\'a}r, P., Girshick, R.: Masked autoencoders are scalable vision learners. pp. 16000--16009. Proceedings of the IEEE/CVF Conference on Computer Vision and Pattern Recognition (CVPR) (2022)

\bibitem{he2016deep}
He, K., Zhang, X., Ren, S., Sun, J.: Deep residual learning for image recognition. pp. 770--778. Proceedings of the IEEE/CVF Conference on Computer Vision and Pattern Recognition (CVPR) (2016)

\bibitem{krishnan2022self}
Krishnan, R., Rajpurkar, P., Topol, E.J.: Self-supervised learning in medicine and healthcare. Nature Biomedical Engineering pp.~1--7 (2022)

\bibitem{li2022covid}
Li, G., Togo, R., Ogawa, T., Haseyama, M.: Covid-19 detection based on self-supervised transfer learning using chest x-ray images. International Journal of Computer Assisted Radiology and Surgery pp.~1--8 (2022)

\bibitem{li2022self}
Li, G., Togo, R., Ogawa, T., Haseyama, M.: Self-knowledge distillation based self-supervised learning for covid-19 detection from chest x-ray images. pp. 1371--1375. Proceedings of the IEEE International Conference on Acoustics, Speech and Signal Processing (ICASSP) (2022)

\bibitem{li2022tri}
Li, G., Togo, R., Ogawa, T., Haseyama, M.: Tribyol: Triplet byol for self-supervised representation learning. pp. 3458--3462. Proceedings of the IEEE International Conference on Acoustics, Speech and Signal Processing (ICASSP) (2022)

\bibitem{liu2021self}
Liu, X., Zhang, F., Hou, Z., Mian, L., Wang, Z., Zhang, J., Tang, J.: Self-supervised learning: Generative or contrastive. IEEE Transactions on Knowledge and Data Engineering  (2021)

\bibitem{loshchilov2019decoupled}
Loshchilov, I., Hutter, F.: Decoupled weight decay regularization. Proceedings of the International Conference on Learning Representations (ICLR) (2019)

\bibitem{noroozi2016unsupervised}
Noroozi, M., Favaro, P.: Unsupervised learning of visual representations by solving jigsaw puzzles. pp. 69--84. Proceedings of the European Conference on Computer Vision (ECCV) (2016)

\bibitem{rahman2021exploring}
Rahman, T., Khandakar, A., Qiblawey, Y., Tahir, A., Kiranyaz, S., Kashem, S.B.A., Islam, M.T., Al~Maadeed, S., Zughaier, S.M., Khan, M.S., Chowdhury, M.E.H.: Exploring the effect of image enhancement techniques on covid-19 detection using chest x-ray images. Computers in Biology and Medicine  \textbf{132},  104319 (2021)

\bibitem{shin2016deep}
Shin, H.C., Roth, H.R., Gao, M., Lu, L., Xu, Z., Nogues, I., Yao, J., Mollura, D., Summers, R.M.: Deep convolutional neural networks for computer-aided detection: Cnn architectures, dataset characteristics and transfer learning. IEEE Transactions on Medical Imaging  \textbf{35}(5),  1285--1298 (2016)

\bibitem{shurrab2022self}
Shurrab, S., Duwairi, R.: Self-supervised learning methods and applications in medical imaging analysis: A survey. PeerJ Computer Science  \textbf{8},  e1045 (2022)

\bibitem{wei2022masked}
Wei, C., Fan, H., Xie, S., Wu, C.Y., Yuille, A., Feichtenhofer, C.: Masked feature prediction for self-supervised visual pre-training. pp. 14668--14678. Proceedings of the IEEE/CVF Conference on Computer Vision and Pattern Recognition (CVPR) (2022)

\bibitem{xie2022simmim}
Xie, Z., Zhang, Z., Cao, Y., Lin, Y., Bao, J., Yao, Z., Dai, Q., Hu, H.: Simmim: A simple framework for masked image modeling. pp. 9653--9663. Proceedings of the IEEE/CVF Conference on Computer Vision and Pattern Recognition (CVPR) (2022)

\end{thebibliography}
\end{document}